\documentclass[conference]{IEEEtran}
\IEEEoverridecommandlockouts
\usepackage{amsmath,amssymb,amsfonts}
\usepackage{algorithmic}
\usepackage{graphicx}
\usepackage{textcomp}
\usepackage{xcolor}
\def\BibTeX{{\rm B\kern-.05em{\sc i\kern-.025em b}\kern-.08em
    T\kern-.1667em\lower.7ex\hbox{E}\kern-.125emX}}

\usepackage[backend=bibtex, style=numeric, sorting=none]{biblatex}
\usepackage[colorlinks]{hyperref}

\begin{document}

\title{An implementation of ROS Autonomous Navigation on Parallax Eddie platform\\
%%{\footnotesize \textsuperscript{*}Note: Sub-titles are not captured in Xplore and should not be used}
%%\thanks{Identify applicable funding agency here. If none, delete this.}
}

\author{\IEEEauthorblockN{1\textsuperscript{st} Hafiq Anas}
\IEEEauthorblockA{\textit{School of Digital Science} \\
\textit{Universiti Brunei Darussalam}\\
Jalan Tungku Link, Brunei \\
hafiq.anas@gmail.com}
\and
\IEEEauthorblockN{2\textsuperscript{nd} Ong Wee Hong}
\IEEEauthorblockA{\textit{School of Digital Science} \\
\textit{Universiti Brunei Darussalam}\\
Jalan Tungku Link, Brunei \\
weehong.ong@ubd.edu.bn}}

\maketitle

\begin{abstract}
This paper presents an implementation of autonomous navigation functionality based on Robot Operating System (ROS) on a wheeled differential drive mobile platform called Eddie robot. ROS is a framework that contains many reusable software stacks as well as visualization and debugging tools that provides an ideal environment for any robotic project development. The main contribution of this paper is the description of the customized hardware and software system setup of Eddie robot to work with an autonomous navigation system in ROS called Navigation Stack and to implement one application use case for autonomous navigation. For this paper, photo taking is chosen to demonstrate a use case of the mobile robot.
\end{abstract}

\begin{IEEEkeywords}
Autonomous Mobile Robot; ROS; Navigation; Odometry; Kinect RGB-D
\end{IEEEkeywords}

\section{Introduction}
The implementation of autonomous navigation based on Robot Operating System
(ROS) on a differential drive mobile platform requires solving problems in robot localization, mapping, and path planning. The usage of ROS in robotic development is becoming increasingly popular due to the robustness and flexibility of the framework. ROS helps to ease the development of any robotic project due to the powerful features provided such as hardware abstraction, message-passing between processes, package management, debugging tools and the availability of packages for various functions of a robotic system \cite{quigley2009ros}. While the desired functionality of autonomous navigation is provided by ROS Navigation stack, successful implementation on an unsupported mobile platform would require proper setup of the stack’s requirements and understanding of its fundamental concepts \cite{quigley2009ros}\cite{eitan}.

A requirement to achieve autonomous navigation is to solve the problem of Simultaneous Localization And Mapping (SLAM) \cite{durrant2006simultaneous}. SLAM is concerned with building a map based on what the robot perceives and localizes itself within that built map \cite{thrun2000probabilistic}. Gmapping algorithm is used for the ROS Navigation Stack’s SLAM functionality \cite{brian}. This is a good option since a good map is able to be generated from SLAM without using too much computational resources.

Generating a suitable traverse path for a mobile robot requires a map of the environment the robot is working in and the generated path should work in such a way that the robot is able to go from one arbitrary position to another without colliding with any obstacles \cite{lumelsky1987path}. A* algorithm is used for ROS Navigation Stack’s Global planner functionality. This approach should enable the Navigation Stack to create an optimal path trajectory for the robot to reach its target location in a shorter period of time.

Odometry is one of the most common localization solutions for Wheeled Mobile Robots (WMRs). The encoders mounted on the wheels can determine a robot’s movement in an environment based on the measured amount and direction of wheel rotation. Eddie robot is equipped with two-wheel encoders \cite{rekleitis2003cooperative}. The implementation of the odometry system is based on interpreting the information from the wheel encoders of the robot to give the robot’s position and orientation information for localization.

SLAM algorithms perform well when a robotic platform is equipped with a high resolution scanning laser rangefinder. However, the cost of such a sensor may not be accessible when the budget is limited. For example, the price of a Hokuyo Laser Sensor and LIDAR sensor costs about \$2,000 and \$75,000 respectively \cite{kneip2009characterization}. An alternative solution is to use an RGB-D sensor like Microsoft Kinect \cite{kneip2009characterization} that costs around \$100 which can emulate a laser scan sensor by using some packages provided in ROS \cite{chad}. Due to its accessibility, Kinect is one of the most commonly used RGB-D sensors for robotic platforms and produces an acceptable map with ROS packages.

In our hardware system setup, we have replaced the original Eddie Control Board with a Raspberry Pi 3, added a Microsoft Kinect RGB-D sensor, a pair of Parallax HB-25 motor controllers, two LM2596 step-down modules, and three 14.4v Lithium Polymer (LiPo) batteries. Further details including the circuit diagram and hardware system diagram will be provided in the next section.

For our software system setup, we have implemented various custom ROS packages to complete the requirements of the Navigation Stack, and this included nodes to process encoder information for odometry localization and nodes to interface between the robot's motor hardware and Navigation Stack. In addition, we have implemented a node that does face detection with an existing Haar Cascade classifier and a node that handles the behavior of the robot which is to detect a person using the aforementioned face detection node and subsequently traverse toward that person to take a photo.

\section{System overview}

\subsection{Hardware framework}

\begin{figure}[!htb]
	\centering
	\includegraphics[width=0.8\linewidth]{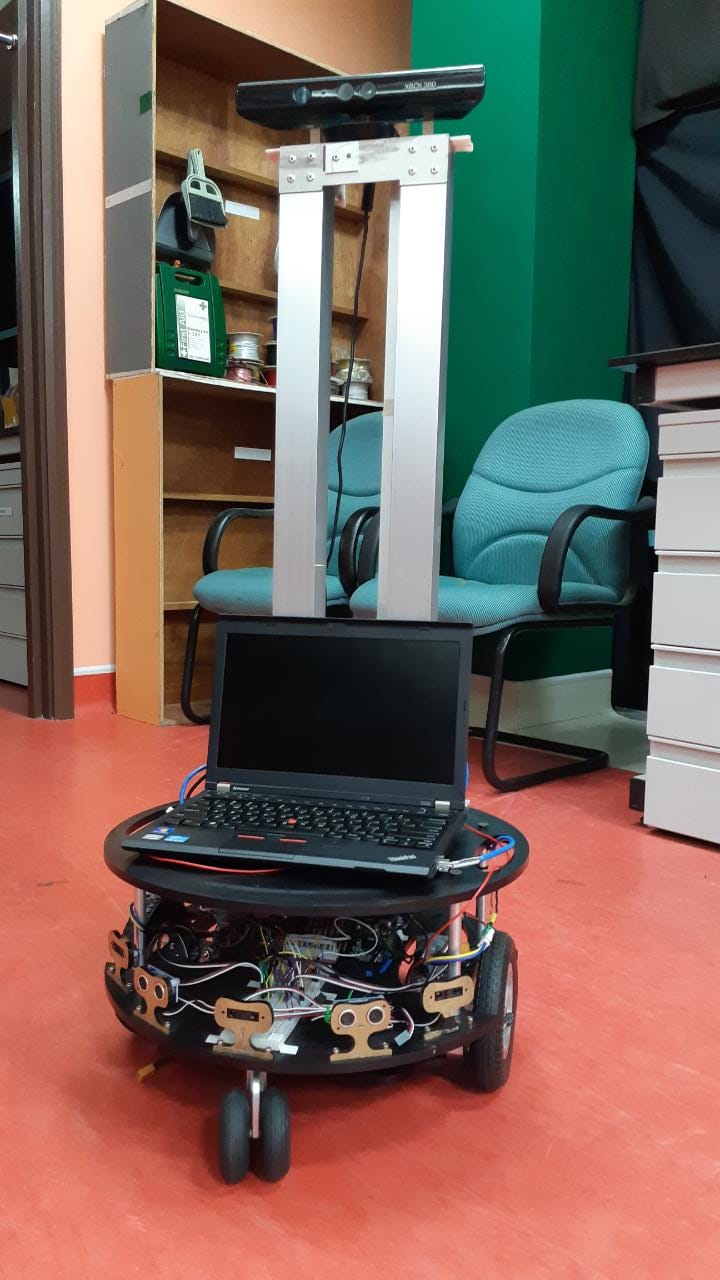}
	\caption{Eddie: A differential drive mobile robot}
	\label{fig:fig1}
\end{figure}

The implementation is carried out on a differential drive wheeled mobile robot platform made by Parallax company, called Eddie robot as shown in \autoref{fig:fig1}. A Raspberry Pi 3 Model B computer and a notebook computer is installed on the robot to run the software system. Additionally, a remote computer is used as the user interface to send instructions to the robot. \autoref{tab:table1} shows the full hardware list of the modified Eddie robot.

\begin{table}[htbp]
	\caption{Eddie hardware specification}
	\label{tab:table1}
	\begin{center}
		\begin{tabular}{|c|c|}
			\hline
			\textbf{Hardware}&\multicolumn{1}{|c|}{\textbf{Details}} \\
			\cline{1-2} 
			\hline
			Computer 1 & Raspberry Pi 3 Model B (Added)\\
			Computer 2 & Lenovo ThinkPad X230 (Added)\\
			Remote computer & Acer Aspire E5 (Added)\\
			Motor controller & 2 * Parallax HB-25 (Added)\\
			Wheel motor & 2 * Parallax motor \#27971\\
			Wheel encoder & 2 * Parallax encoder \#29321\\
			RGB-D sensor & Microsoft Kinect v1 (Added)\\
			Step down module & 2 * LM2596 module (5V, 12V) (Added)\\
			Power supply & 3 * 14.4v 5000mAH LiPo (Added)\\
			\hline
		\end{tabular}
	\end{center}
\end{table}

\subsubsection{Eddie robot system hardware setup}
\begin{figure}[!htb]
	\centering
	\includegraphics[width=1.0\linewidth]{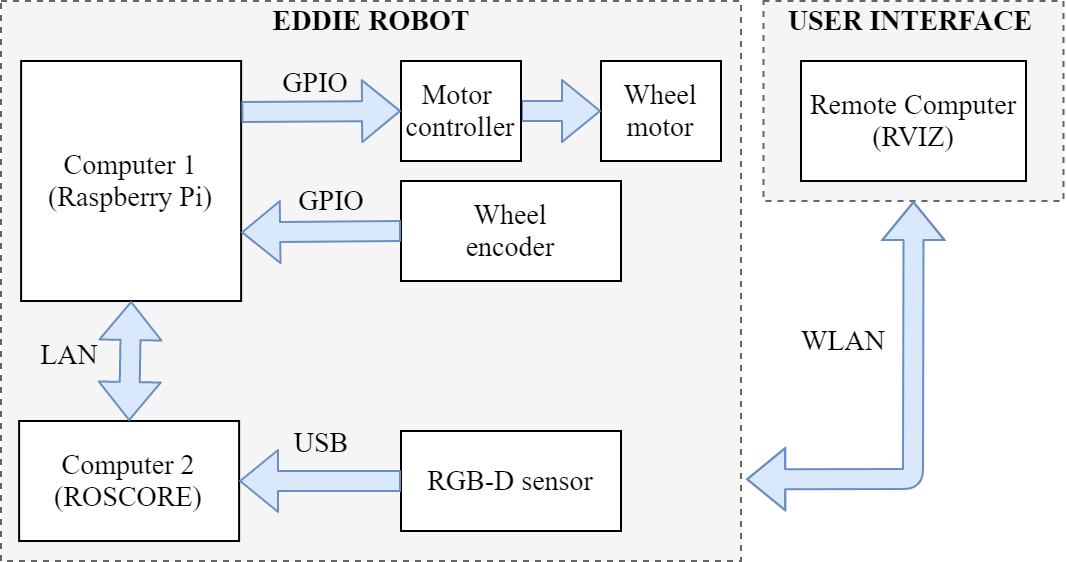}
	\caption{Hardware system of Eddie robot}
	\label{fig:fig2}
\end{figure}

\autoref{fig:fig2} shows the block diagram of the hardware setup of the robot. Several modifications have been done on the Eddie robot to implement ROS Navigation Stack. The existing control board has been replaced by a Raspberry Pi 3 Model B computer which is used to interface with the sensors and motors. An aluminum frame is installed on the platform with a Microsoft Kinect v1 fitted on top and connected to a notebook computer through a USB port. Due to the low computational power of Raspberry Pi, a notebook computer is used to run the more computationally demanding functionality such as SLAM and navigation. Additionally, this notebook computer communicates with the Raspberry Pi through a direct LAN link. A remote computer is connected to the robot through SSH over WLAN and is used to run a visualization software called RVIZ to monitor operations and send motor commands. The hardware specifications of all the computers in this setup is shown in \autoref{tab:table2}.

\begin{figure*}[!htb]
	\centering
	\includegraphics[width=0.9 \linewidth]{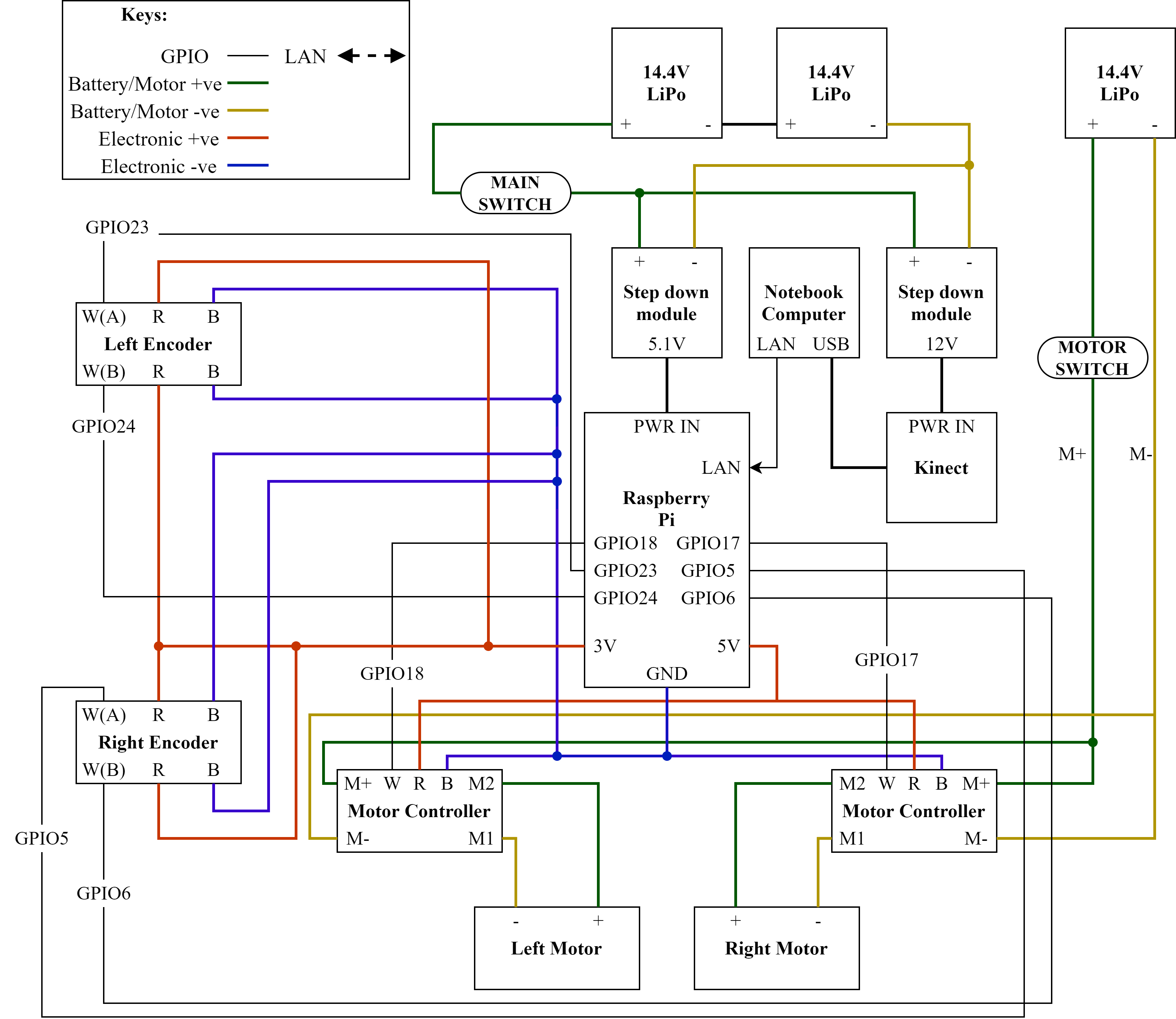}
	\caption{Wiring diagram of Eddie robot}
	\label{fig:fig3}
\end{figure*}

The wiring diagram of the hardware setup is shown in \autoref{fig:fig3}. The Raspberry Pi is connected to all GPIO supported hardware modules (i.e wheel encoders and motor controllers), and the DC Power ports (i.e 3V, 5V) are used to supply voltage for these modules. Since the Raspberry Pi and Kinect require different voltages, two 14.4V Lithium Polymer (LiPo) batteries are connected in series and a step-down module is used to lower down the combined voltage of 28.8V to 5.1V and 12V for the two devices respectively. In addition, another identical LiPo battery is used to supply voltage for the wheel motors. Different batteries are used to supply the electronic circuits and the motors to avoid the running motors to interfere with the voltage of the electronic circuits. The Kinect is connected to the Notebook instead of the Raspberry Pi since the Notebook has a higher computational capability and a LAN cable is connected to allow communications between the two computers in our software system setup.

\begin{table}[htbp]
	\caption{Computer hardware specifications}
	\label{tab:table2}
	\begin{center}
		\begin{tabular}{|c|c|c|c|}
			\hline
			\textbf{Computer}&\multicolumn{3}{|c|}{\textbf{Computers}} \\
			\cline{2-4} 
			\textbf{Parts} & \textbf{\textit{Raspberry Pi}}& \textbf{\textit{Notebook}}& \textbf{\textit{Remote}} \\
			\hline
			CPU & Cortex-A53 & i5-3320M & i5-5200U\\
			GPU & VideoCore IV & Intel HD & Nvidia 840M\\
			RAM & 1GB & 4GB & 8GB\\
			STORAGE & 16GB & 500GB & 500GB\\
			OS & Raspbian Stretch & Ubuntu 18.04 & Ubuntu 18.04\\
			ROS & Kinetic & Melodic & Melodic\\
			\hline
		\end{tabular}
	\end{center}
\end{table}

\subsection{Software framework}

The implementation of autonomous navigation on Eddie robot uses ROS as a framework to provide a network environment where information from the Raspberry Pi and Notebook can be transmitted from one to another. The software framework is mainly divided into two sections which are: navigation system and photo-taking behavior. Several ROS nodes have been implemented specifically for the Eddie robot's hardware listed in \autoref{tab:table1} to enable the Navigation stack to work with the current hardware setup. Additionally, the parameters of the Navigation stack have been optimized through trial and error. A photo-taking behavior has been implemented to demonstrate a use case for the navigation system. A detailed explanation regarding the implementation of the navigation system will be given in \autoref{Section:sec3}.

\subsubsection{Software structure of navigation system}
\begin{figure}[!htb]
	\centering
	\includegraphics[width=1.0\linewidth]{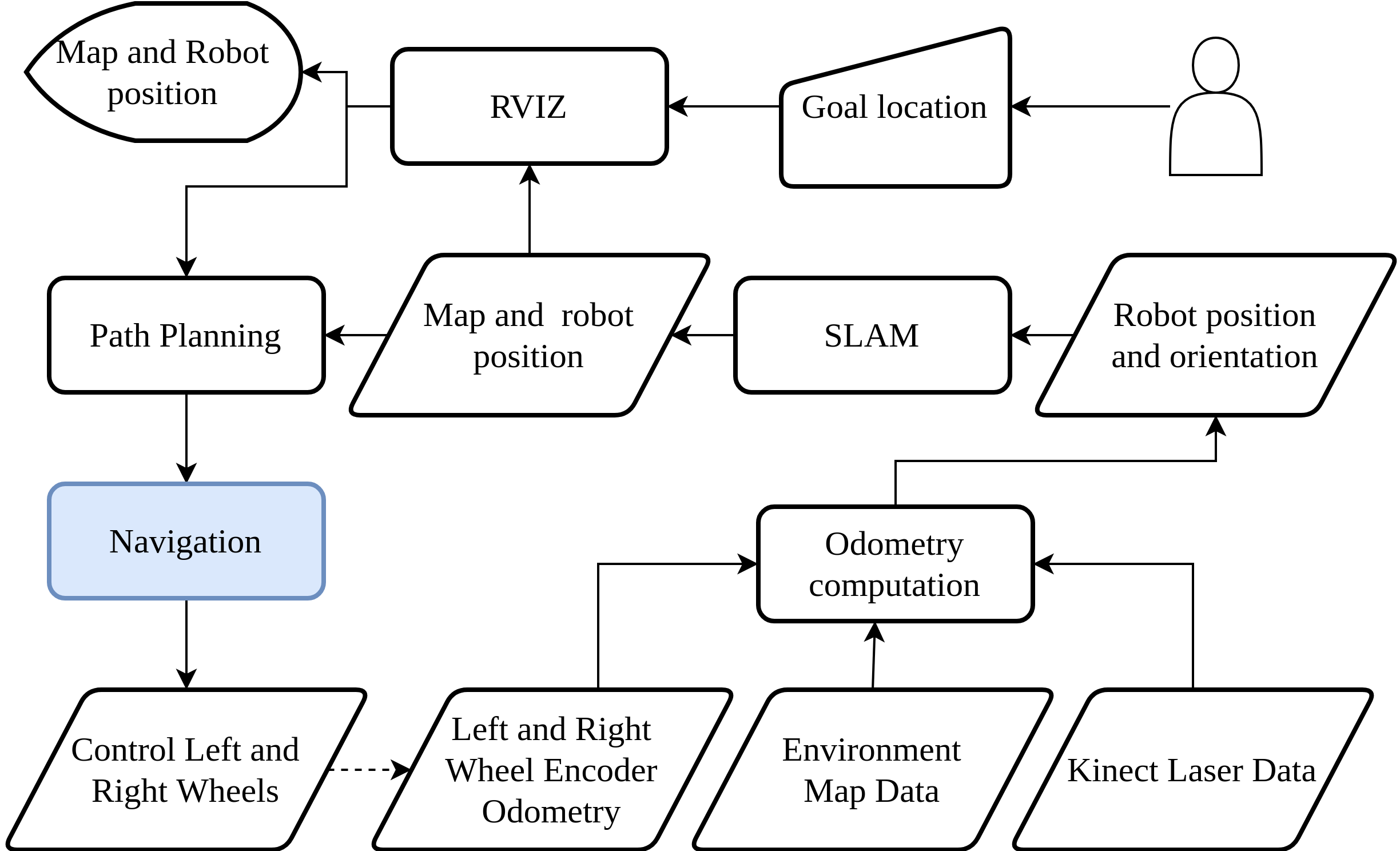}
	\caption{Navigation system software flowchart}
	\label{fig:fig4}
\end{figure}

The flowchart of the navigation system is shown in \autoref{fig:fig4}. Initially, a user sends a goal location containing position and orientation information to the navigation system through the interface of the ROS visualization package, RVIZ. Then, the system starts to generate a suitable path of trajectory and sends motor commands (i.e linear and angular velocity) to the two-wheel motor drivers. While the robot is moving, the odometry is continuously updated and used by SLAM for localization and mapping. Finally, the map which was generated by SLAM and robot location (pose) can be visualized on RVIZ. Additionally, teleoperation allows users to assume manual remote control of the robot from a remote computer and this was used frequently during the experimentation phase. Detailed information on the implementation of ROS is given in \autoref{Section:sec3}.

\subsubsection{Software structure of photo taking behavior}
\begin{figure}[!htb]
	\centering
	\includegraphics[width=1.0\linewidth]{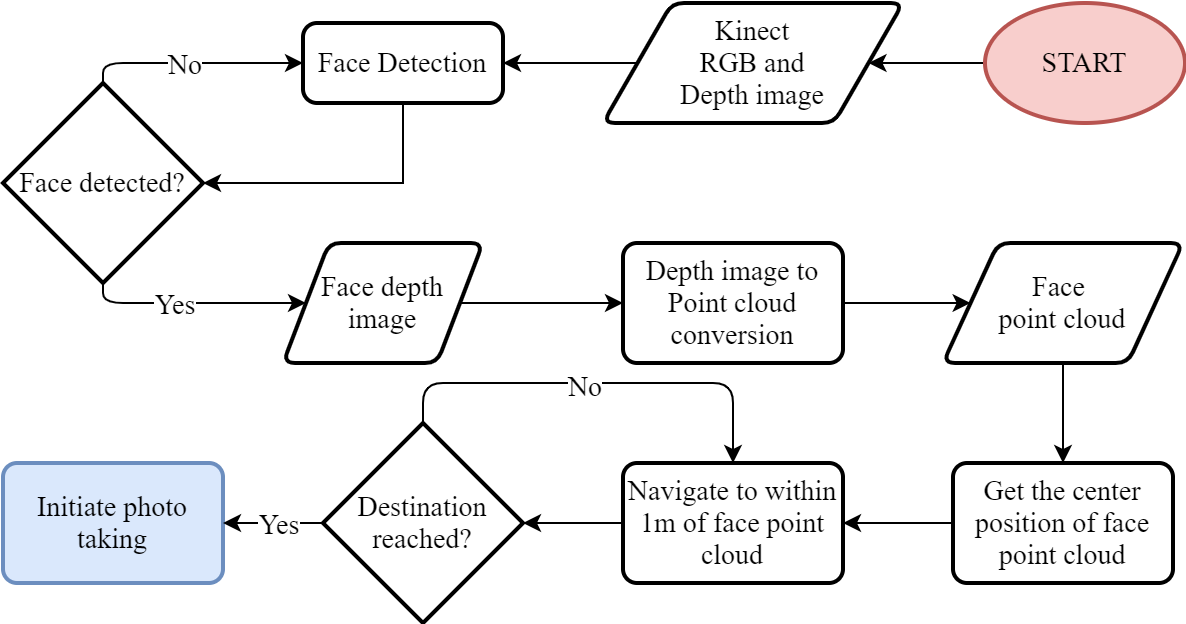}
	\caption{Photo taking behavior flow flowchart}
	\label{fig:fig5}
\end{figure}

The flowchart of the photo-taking system is shown in \autoref{fig:fig5}. Firstly, the live raw image data from the Kinect is used by the face detector to detect faces. Then, the detected face contains depth information which is converted to point cloud data and this is used to determine the position of the human. Finally, the human location will be used as a goal position and sent to the navigation system where the robot will start its autonomous navigation toward the human.

After successful navigation, the OpenCV library is used to save and show the taken picture. A more detailed discussion is provided in \autoref{Section:sec4}.

\section{Implementation of navigation system}
\label{Section:sec3}
\begin{figure}[!htb]
	\centering
	\includegraphics[width=1.0\linewidth]{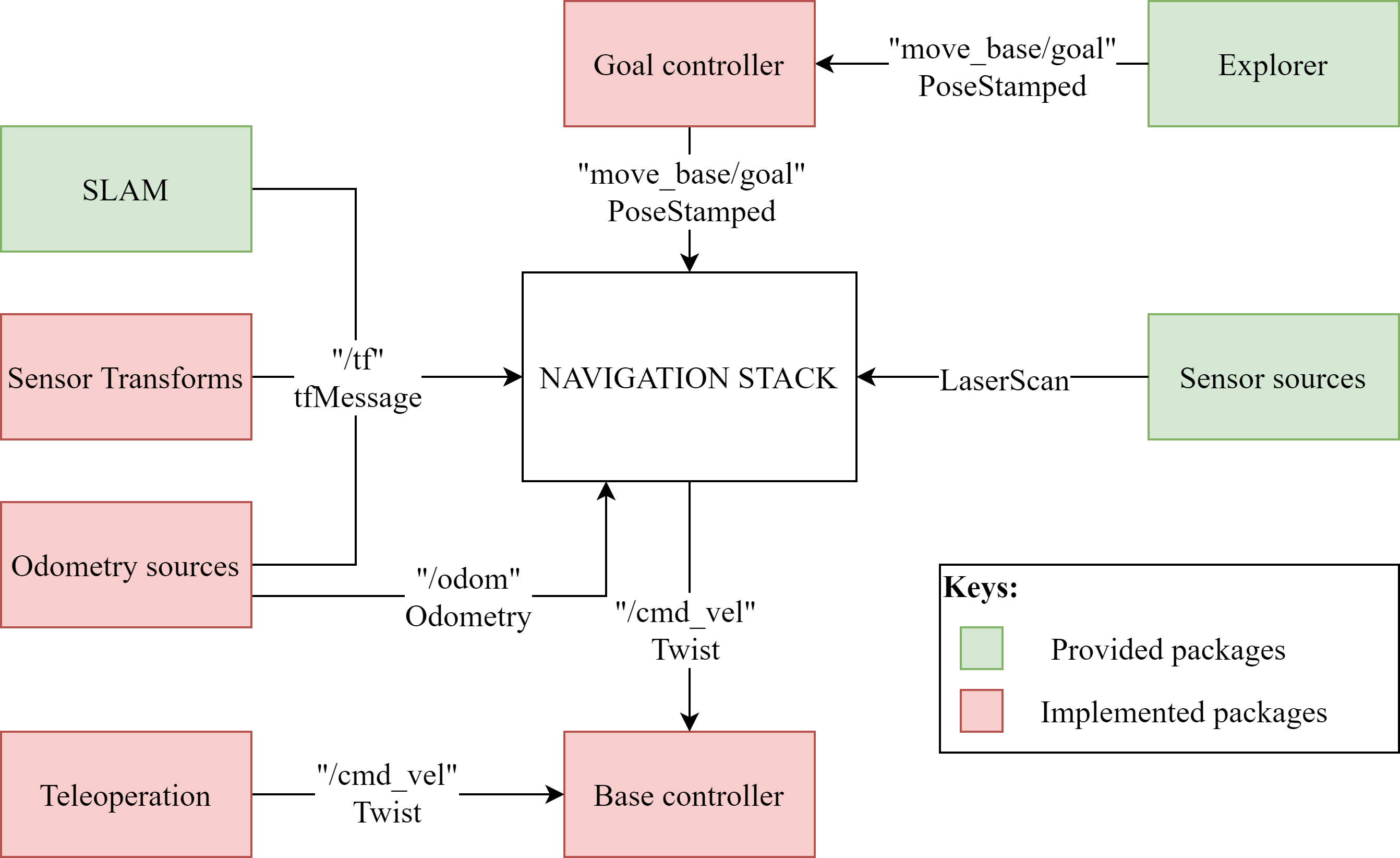}
	\caption{ROS Navigation stack setup}
	\label{fig:fig6}
\end{figure}

The implementation process of autonomous navigation will be based on satisfying the requirements to run ROS Navigation stack \cite{eitan}. \autoref{fig:fig6} shows the packages that are required for the Navigation stack to run. Navigation stack, SLAM, Sensor sources, and Explorer are packages obtained from the ROS library while the remaining are packages implemented in this paper. 

The odometry sources package provides odometry information computed from the two wheel encoders. The Sensor sources package provides Laser scans information from the Kinect sensor. Sensor Transforms package provides the necessary physical position and orientation relationship between different robot components such as robot base, wheel position, and Kinect sensor. Altogether, these packages are used by SLAM to properly localize the robot based on wheel encoder and Kinect sensor information. The Teleoperation package gets keyboard inputs from the remote computer to manually drive the robot. The Base controller package accepts velocity commands from either the Navigation stack or the Teleoperation package that interprets the velocity commands and sends appropriate signals to the robot motors to drive the robot. The goal controller package accepts goal location input from RVIZ and finally, the Explorer package sends the goal location to the Navigation stack based on the unexplored regions of the map generated by SLAM that allows the robot to explore its surroundings.

The overall overview and details of the Navigation Stack packages' implementation will be explained in \autoref{Section:sec3d}. Meanwhile, the following sections will be about the implementations of the robot's motor control, wheel encoder odometry, base controller, base teleoperator, goal controller, and photo-taking behavior.

The Teleoperation, Goal Controller, and Base Controller packages shown in \autoref{fig:fig6} require implementing a robot motor control (\autoref{Section:sec3a}) which converts velocity commands into motor commands to drive the robot. While the Odometry Sources package requires implementing a robot encoder reader (\autoref{Section:sec3b}) and odometry model (\autoref{Section:sec3c}) for robot localization using encoder hardware.

\subsection{Motor control model}
\label{Section:sec3a}
PIGPIO library \cite{croston_2019} is used to control the wheel motor controllers with Pulse Width Modulation (PWM). A relationship between velocity and Pulse Width (PW) of the left and right wheel motors is defined. The relationship between the PWs of the motor controllers and velocity is assumed to be proportional \cite{dewangan2012pwm}.

Firstly, the PWs of the motor controllers are set up in such a way that the two wheels turn at maximum speed and a tachometer is used to measure the Revolution Per Minute (RPM) of the two wheels at that speed. The linear equation to establish the relationship between RPM and PW (time in milliseconds) is given below:

\begin{equation}
\label{eqn:eq1}
RPM = K \cdot PW
\end{equation}	

Where $K$ is the proportionality constant between $RPM$ and $PW$. In \autoref{eqn:eq1}, $K$ can be obtained since $PW$ and $RPM$ are known. By using the robot's wheel radius $r$, an equation between $PW$ and linear velocity $V$ is derived as below:

\begin{equation}
\label{eqn:eq2}
V = \frac{2{\pi}r}{60}\cdot{RPM}
\end{equation}

\begin{equation}
\label{eqn:eq3}
V = \frac{2{\pi}r}{60}\cdot{K}\cdot{PW}
\end{equation}

Based on \autoref{eqn:eq3}, a program is implemented to convert velocity commands into motor commands to control the speed of the wheel motors.

\subsection{Encoder reading model}
\label{Section:sec3b}
\begin{figure}[!htb]
	\centering
	\includegraphics[width=1.0\linewidth]{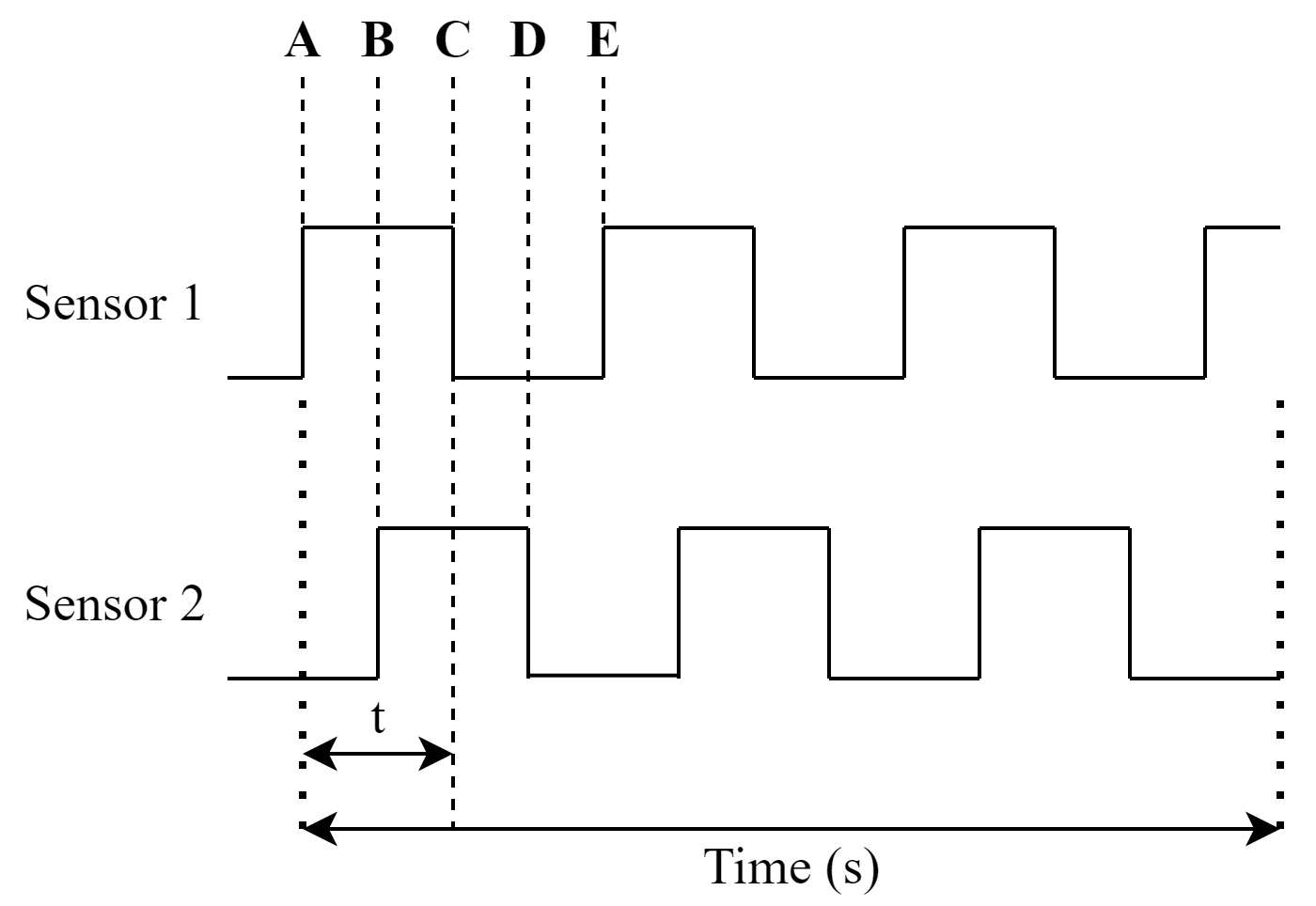}
	\caption{Visualizaion of signals generated by the wheel encoders}
	\label{fig:fig7}
\end{figure}

RPI.GPIO library \cite{fivdi_2020} is used to detect generated pulses from the encoders for localization purpose. Two wheel encoders are used to deduce the speed and travel direction by measuring the number of pulses registered from each wheel. Each wheel encoder has two sensors and is capable of registering a distance resolution of 1/36th of the robot's wheel circumference. \autoref{fig:fig7} \cite{parallaxmotormount} shows the signals generated by the encoders.

Knowing that the sensor has 1/36th resolution of wheel circumference, the sensor produces 36 pulses for every complete revolution of the wheel. Based on this, the distance traveled in the duration of one pulse is given below:

\begin{equation}
\label{eqn:eq4}
d = \frac{2{\pi}r}{36}
\end{equation}

Where $d$ is the distance travelled in meters and $r$ is the radius of the wheel in meters.\\

Speed information is computed in \autoref{eqn:eq5}. 

\begin{equation}
\label{eqn:eq5}
s = \frac{d}{t}
\end{equation}

Where $t$ is the Pulse Width (PW) which is the time that the encoder's signal stays high (i.e A-C) as shown in \autoref{fig:fig7}.\\

\subsection{Odometry model}
\label{Section:sec3c}
This model is used to describe the robot's position and orientation as a function of the movement of its wheels but ignores any physics involved in making that motion. So, it is used to estimate a robot's position using the sensor information from wheel encoders and this is known as dead reckoning \cite{inbook}.

\begin{figure}[!htb]
	\centering
	\includegraphics[width=0.9\linewidth]{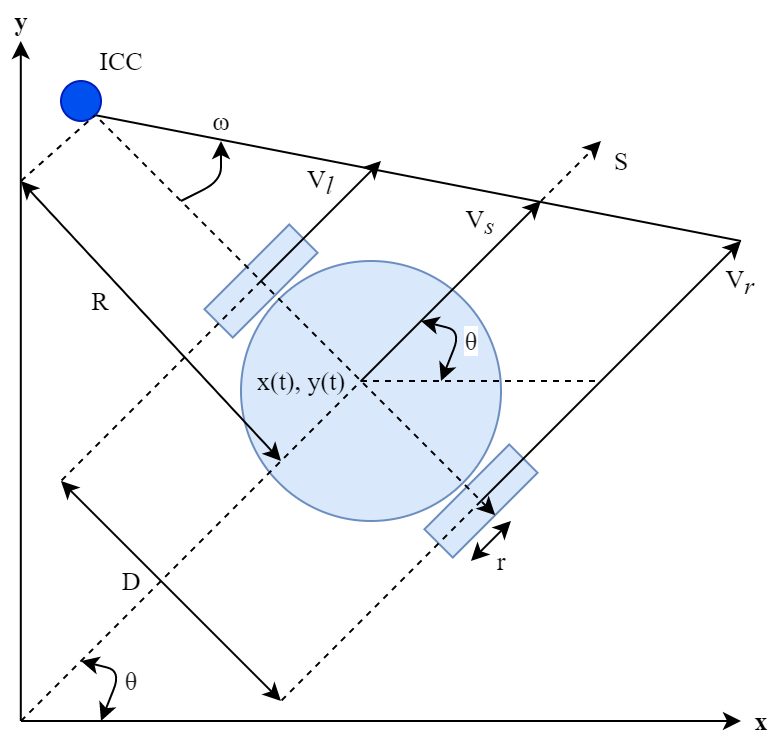}
	\caption{Robot Cartesian coordinate system and variable definitions}
	\label{fig:fig8}
\end{figure}

In this case, the robot has two independent controllable wheels which are mounted on a common axis and the control of its movement direction is by manipulating the difference in velocities of its two wheels. Under the assumption that the wheels are fixed on the robot, the two wheels must describe arcs on the plane such that the robot rotates around a point known as ICC (Instantaneous Center of Curvature) which lies on the wheels' common axis (\autoref{fig:fig8}) \cite{inbook}. The X-Y axes refer to the global reference frame in the environment.

Let us consider the contribution of each wheel’s linear velocity to the movement direction of the robot.

\begin{itemize}
	\item When $V_l$ and $V_r$ are equal, the robot moves in a straight line.
	\item When $V_l$ and $V_r$ are not equal, the robot moves in the direction of wheel with lower linear velocity.
	\item When $V_l$ and $V_r$ are equal and in opposite direction, the robot spins at stationary position. If the left wheel is in forward direction, the robot spins clockwise; and vice versa.
\end{itemize}

If the linear velocity of the left and right wheels are ${V}_l$ and ${V}_r$ respectively, and the right and left wheels are separated by distance $D$ in meters, then linear velocities can be computed as follows:

\begin{equation}
\label{eqn:eq6}
V_l = (R - \frac{D}{2}) \cdot {\omega}
\end{equation}
\begin{equation}
\label{eqn:eq7}
V_r = (R + \frac{D}{2}) \cdot {\omega}
\end{equation}

Where, $R$ is the distance between the center of the robot to the ICC in meters and $\omega$ is the rate of rotation about the ICC in radian per second. The following equations (\autoref{eqn:eq8} and \autoref{eqn:eq9}) are derived by rearranging \autoref{eqn:eq6} and \autoref{eqn:eq7} to determine $R$ and $\omega$ at any instance of time.

\begin{equation}
\label{eqn:eq8}
{\omega} = \frac{({V}_r - {V}_l)}{D}
\end{equation}
\begin{equation}
\label{eqn:eq9}
R = \frac{D}{2} \cdot \frac{{{V}_r + {V}_l}}{{V}_l - {V}_r}
\end{equation}

Thus, the instantaneous velocity at the center point of the robot is given by \autoref{eqn:eq10} and can be derived as \autoref{eqn:eq11} from \autoref{eqn:eq8} and \autoref{eqn:eq9}.

\begin{equation}
\label{eqn:eq10}
{V}_s = \frac{V_l + V_r}{2}
\end{equation}

\begin{equation}
\label{eqn:eq11}
{V}_s = {\omega}{R}
\end{equation}

Using the center point between the wheels as the origin of the robot, and writing $\theta$ as the orientation of the robot with respect to the x-axis of a global Cartesian coordinate system shown in \autoref{fig:fig8}, a set of motion equations for the robot can be generated as shown below in \autoref{eqn:eq12} and \autoref{eqn:eq13} where ${V}_x$ and ${V}_y$ are the velocity in x and y direction respectively. These equations can be rearranged to obtain \autoref{eqn:eq14} and \autoref{eqn:eq15}.

\begin{equation}
\label{eqn:eq12}
\cos\theta = \frac{V_x}{V_s}
\end{equation}
\begin{equation}
\label{eqn:eq13}
\sin\theta = \frac{V_y}{V_s}
\end{equation}
\begin{equation}
\label{eqn:eq14}
V_x = V_s \cdot \cos\theta
\end{equation}
\begin{equation}
\label{eqn:eq15}
V_y = V_s \cdot \sin\theta
\end{equation}

The movement in x and y direction as well as the angular movement are obtained by taking the integration of the respective direction components at small time steps, we can simplify the computation of change in $x(t)$, $y(t)$ and $\theta(t)$ as below:

\begin{equation}
\label{eqn:eq16}
\delta{x(t)} = {\int_0^t V_s(t) \cdot \cos[\theta(t)] \cdot \delta{t}}
\end{equation}
\begin{equation}
\label{eqn:eq17}
\delta{y(t)} = {\int_0^t V_s(t) \cdot \sin[\theta(t)] \cdot \delta{t}}
\end{equation}
\begin{equation}
\label{eqn:eq18}
\delta{\theta(t)} = {\int_0^t \omega(t) \cdot \delta{t}}
\end{equation}

Where $\delta{t}$ is time step where change of position in $x$, $y$ and orientation $\theta$ is computed. In this case, the time step can be based on the Pulse Width (PW) from the wheel encoders' readings that is the period when the encoder signal is detected as high (see \autoref{Section:sec3b}).

Suppose that the robot is positioned at ($x$, $y$, $\theta$) at time $t$ in the coordinate system (\autoref{fig:fig8}), then the next position ($x'$, $y'$, $\phi'$) can be calculated using the equations below:

\begin{equation}
x' = x + \delta{x(t)}
\end{equation}
\begin{equation}
y' = y + \delta{y(t)}
\end{equation}
\begin{equation}
\theta' = \theta + \delta{\theta(t)}
\end{equation}
\begin{equation}
x' = x + {\int_0^t V_s(t) \cdot \cos[\theta(t)] \cdot \delta{t}}
\end{equation}
\begin{equation}
y' = y + {\int_0^t V_s(t) \cdot \sin[\theta(t)] \cdot \delta{t}}
\end{equation}
\begin{equation}
\theta{'} = \theta + {\int_0^t \omega(t) \cdot \delta{t}}
\end{equation}

Where, [$x$, $y$, $\theta$] is the current position and orientation of the robot and [$\delta{x(t)}$, $\delta{y(t)}$, $\delta{\theta(t)}$] is the change in position and orientation at time step $t$. So, this Odometry model is used to estimate the position and orientation of the robot for localization.

\subsection{Navigation stack setup}
\label{Section:sec3d}
\autoref{fig:fig6} shows the complete system for the Navigation stack \cite{quigley2009ros}\cite{eitan} and the packages that are required to run the stack. \autoref{tab:table3} shows the list of packages that were implemented or obtained from the ROS library. This section will mainly discuss the tuning of the Navigation stack and some of its packages, and the implementation of custom packages. The implementation of Face Detector and Photo Taking Behavior packages will be discussed in \autoref{Section:sec4}

\begin{table}[htbp]
	\caption{ROS Navigation stack packages}
	\label{tab:table3}
	\begin{center}
		\begin{tabular}{|c|c|}
			\hline
			\textbf{Packages}&\multicolumn{1}{|c|}{\textbf{Status}} \\
			\cline{1-2} 
			\hline
			ODOMETRY SOURCES & Implemented\\
			GOAL CONTROLLER & Implemented\\
			BASE CONTROLLER & Implemented\\
			TELEOPERATION & Implemented\\
			SENSOR TRANSFORMS & Implemented\\
			SLAM & ROS library\\
			SENSOR SOURCES & ROS library\\
			EXPLORER & ROS library\\
			\hline
		\end{tabular}
	\end{center}
\end{table}

For reference, the SLAM and EXPLORER packages used are called "gmapping" \cite{brian} and "explorer{\textunderscore}lite" \cite{jim} respectively. Meanwhile, there are two packages used in SENSOR SOURCES which are "openni{\textunderscore}launch" \cite{patrickjames} and "depthimage{\textunderscore}to{\textunderscore}laserscan" \cite{chad} respectively. 

ODOMETRY SOURCES package is used to provide localization information based on the robot's wheel encoders, GOAL CONTROLLER package is used to send a goal location to the Navigation Stack for the robot to move to and is used in the implementation of photo-taking behavior. BASE CONTROLLER package is used to drive the robot after receiving drive commands from the Navigation Stack. TELEOPERATION package is used to remotely control the robot through keyboard inputs for debugging purposes. SLAM package is used to create a map based on the robot's sensor readings from the ODOMETRY SOURCES and SENSOR SOURCES packages. SENSOR SOURCES is used to send the robot's Laser Scan information from its Kinect sensor. Finally, EXPLORER package is used for the robot to explore the environment by sending goal locations to the Navigation Stack based on unexplored regions of the environment.\\

\subsubsection{Sensor Transforms package}
\begin{figure}[!htb]
	\centering
	\includegraphics[width=1.0\linewidth]{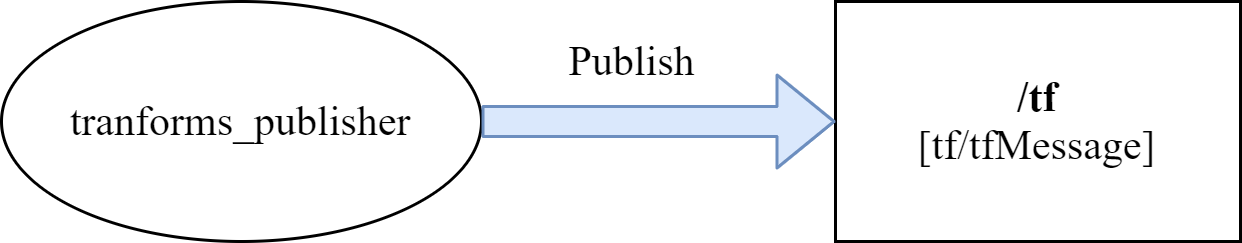}
	\caption{robot\textunderscore{transforms} package node}
	\label{fig:fig9}
\end{figure}

A package called \textbf{robot{\textunderscore}transforms} has been created which maintains a Transforms (TF) tree for the robot. Within the package, a node called \textit{transforms{\textunderscore}publisher} publishes messages on a topic named ``/tf" with message type ``tf/tfMessage" as shown in \autoref{fig:fig9}.

A TF tree is required to maintain the position and orientation relationship between different sensors attached to a robot platform. In this case, the reference point of Odometry is at the center of the robot base. So, a TF between the Odometry (Odom), robot base (base{\textunderscore}link), and Kinect sensor (camera{\textunderscore}link) is established using TF library in ROS. \autoref{tab:table4} shows the 3D measurements of the robot's TF tree. Note that the Kinect sensor (camera{\textunderscore}link) is installed at the height of 1 meter above the center of the robot base (base{\textunderscore}link).

\begin{table}[htbp]
	\caption{Robot Transforms tree measurements}
	\label{tab:table4}
	\begin{center}
		\begin{tabular}{|c|c|}
			\hline
			\textbf{TF}&\multicolumn{1}{|c|}{\textbf{x, y, z (in m)}} \\
			\cline{1-2} 
			\hline
			camera{\textunderscore}link & 0.0, 0.0, 1.0\\
			base{\textunderscore}link & 0.0, 0.0, 0.0\\
			odom & 0.0, 0.0, 0.0\\
			\hline
		\end{tabular}
	\end{center}
\end{table}

\subsubsection{Odometry Source package}
\begin{figure}[!htb]
	\centering
	\includegraphics[width=1.0\linewidth]{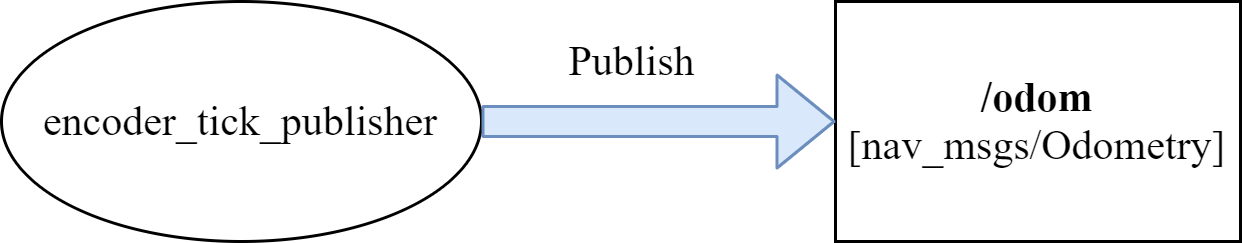}
	\caption{robot{\textunderscore}odometry package node}
	\label{fig:fig10}
\end{figure}

An Odometry source is required to provide position($x$, $y$) and orientation ($\theta$) information to the Navigation stack. The odometry information is referenced from the center of the robot base and computed based on the signal detected from the wheel encoders. A package called \textbf{robot{\textunderscore}odometry} has been created which performs robot localization and the implementation is based on Encoder Reading model and Odometry model (\autoref{Section:sec3b} and \autoref{Section:sec3c} respectively). Within the package, a node called \textit{encoder{\textunderscore}tick{\textunderscore}publisher} publishes messages on a topic named ``/odom" with type ``nav{\textunderscore}msgs/Odometry" as shown in \autoref{fig:fig10}, which encapsulate the pose (x, y) and orientation ($\theta$) of the robot.

\subsubsection{Base Controller package}
\begin{figure}[!htb]
	\centering
	\includegraphics[width=1.0\linewidth]{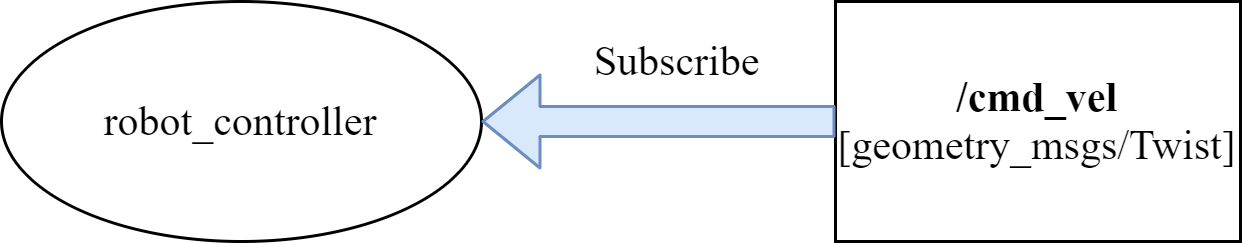}
	\caption{robot{\textunderscore}controller package node}
	\label{fig:fig11}
\end{figure}

A package called \textbf{robot{\textunderscore}controller} has been created which controls the robot and the implementation is based on the Motor Control layer (\autoref{Section:sec3a}). Within the package, a node called \textit{robot{\textunderscore}controller} subscribes to a topic named ``/cmd{\textunderscore}vel" which publishes messages with type ``geometry{\textunderscore}msgs/Twist" as shown in \autoref{fig:fig11}. The message type geometry{\textunderscore}msgs/Twist stores velocity commands in the form of linear and angular velocities.

Since path planning is done within the Navigation stack (\autoref{fig:fig4}), the Navigation stack is responsible for publishing velocity commands on /cmd{\textunderscore}vel topic and this package converts velocity commands into motor control commands that drive the robot.

\subsubsection{Teleoperation package}
\begin{figure}[!htb]
	\centering
	\includegraphics[width=1.0\linewidth]{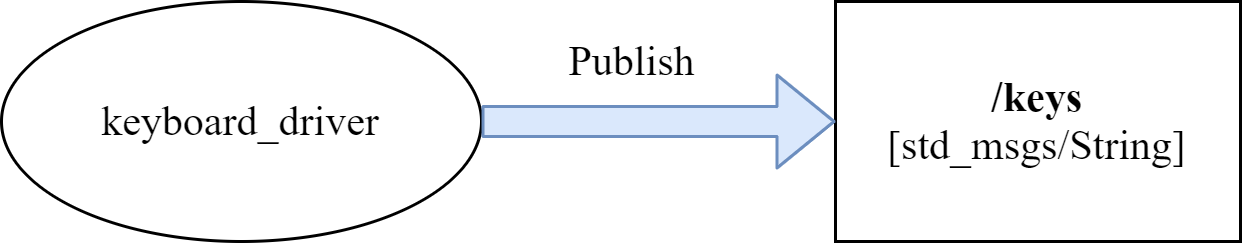}
	\caption{Teleopration package node}
	\label{fig:fig12}
\end{figure}

\begin{figure}[!htb]
	\centering
	\includegraphics[width=0.9\linewidth]{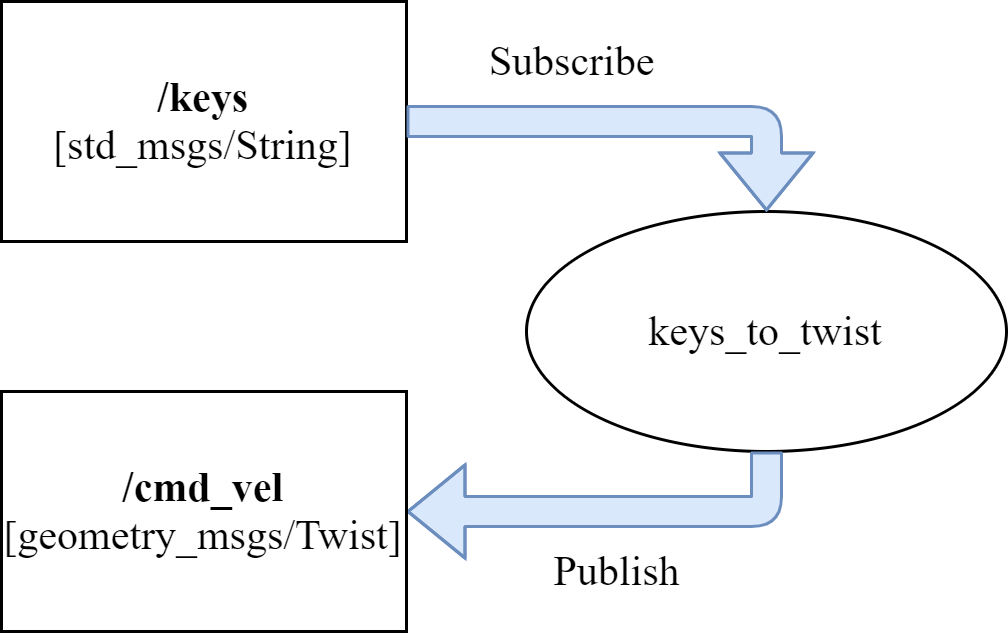}
	\caption{Teleoperation package node}
	\label{fig:fig13}
\end{figure}

Within the \textbf{robot{\textunderscore}controller} package, two nodes called \textit{keyboard{\textunderscore}driver} (\autoref{fig:fig12}) and \textit{keys{\textunderscore}to{\textunderscore}twist} (\autoref{fig:fig13}) have been created which detect keyboard key presses on a remote computer and convert the key presses into velocity commands respectively. ``keyboard{\textunderscore}driver" publishes messages on a topic named ``/keys" with type ``std{\textunderscore}msgs/String" while ``keys{\textunderscore}to{\textunderscore}twist" subscribes to the topic named ``/keys" and publishes messages with type ``geometry{\textunderscore}msgs/Twist".

The two nodes mentioned above is used to publish velocity commands for the \textbf{robot{\textunderscore}controller} (\autoref{fig:fig11}) to manually move the robot using key presses from a remote computer. The use of this functionality was mainly for testing and debugging of the Odometry system as well as for potential remote control applications. So, the \textit{robot{\textunderscore}controller} node can subscribe to the topic ``/cmd{\textunderscore}vel" published by the Navigation stack or \textit{Teleoperation} node.

\subsubsection{Goal controller package}
\begin{figure}[!htb]
	\centering
	\includegraphics[width=1.0\linewidth]{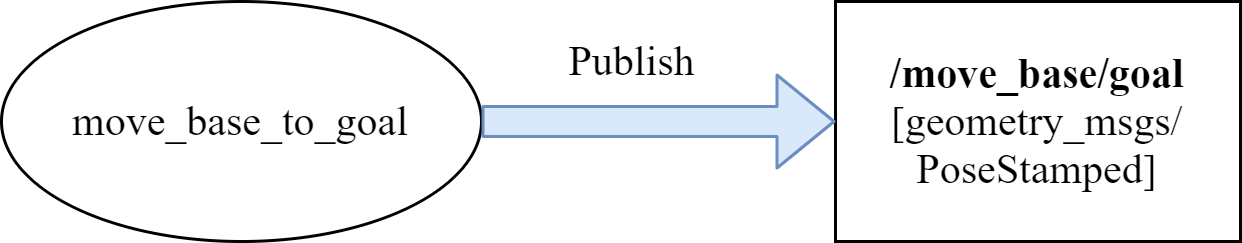}
	\caption{robot{\textunderscore}send{\textunderscore}goals package node}
	\label{fig:fig14}
\end{figure}

A package called \textbf{robot{\textunderscore}send{\textunderscore}goals} has been created which sends goal commands to the Navigation stack without the RVIZ graphical interface. Within the package, a node called \textit{move{\textunderscore}base{\textunderscore to\textunderscore}goal} shown in \autoref{fig:fig14} sends a goal message with a message type of ``geometry{\textunderscore}msgs/PoseStamped" that contains the position ($x$, $y$) and orientation($\theta$) of the goal location to the Navigation stack. A goal location input in the form of x, y and orientation $\theta$ can be passed from a function that intends to move the robot to a specific location.

Besides being used by the Navigation stack, this package is also used in the implementation of photo-taking behavior (\autoref{Section:sec4}) where the position and orientation of a human face are set as a goal position where the robot will go to for photo taking.

\subsection{Navigation stack parameter tuning}
The default parameters in the configuration files of the ROS Navigation stack did not work well for the robot as it moved erratically with highly fluctuating ground wheel speed and did not manage to stop at the designated goal position. So, custom parameters built for the robot hardware specification are required to run the Navigation stack successfully. Through visual inspection, these parameters were found to provide the best balance between navigation performance and computational cost. \autoref{tab:table5}, \autoref{tab:table6}, \autoref{tab:table7} and \autoref{tab:table8} show the tuned parameters. The tuned parameters and detailed explanation can be found in \cite{zheng2017ros}.

\begin{table}[htbp]
	\caption{Base local planner parameters}
	\label{tab:table5}
	\begin{center}
		\begin{tabular}{|c|c|}
			\hline
			\textbf{Parameters}&\multicolumn{1}{|c|}{\textbf{Values}} \\
			\cline{1-2} 
			\hline
			controller{\textunderscore}frequency & 15.0\\
			holonomic{\textunderscore}robot & false\\
			yaw{\textunderscore}goal{\textunderscore}tolerance & 0.2\\
			xy{\textunderscore}goal{\textunderscore}tolerance & 0.3\\
			sim{\textunderscore}time & 4.0\\
			vtheta{\textunderscore}samples & 40\\
			\hline
		\end{tabular}
	\end{center}
\end{table}

\begin{table}[htbp]
	\caption{Costmap common parameters}
	\label{tab:table6}
	\begin{center}
		\begin{tabular}{|c|c|}
			\hline
			\textbf{Parameters}&\multicolumn{1}{|c|}{\textbf{Values}} \\
			\cline{1-2} 
			\hline
			footprint & [[-0.1, -0.1], [-0.1, 0.1], [0.1, 0.1], [0.1,-0.1]]\\
			transform{\textunderscore}tolerance & 0.3\\
			publish{\textunderscore}frequency & 10.0\\
			inflation{\textunderscore}radius & 1.75\\
			cost{\textunderscore}scaling{\textunderscore}factor & 2.58\\
			\hline
		\end{tabular}
	\end{center}
\end{table}

\begin{table}[htbp]
	\caption{Global costmap parameters}
	\label{tab:table7}
	\begin{center}
		\begin{tabular}{|c|c|}
			\hline
			\textbf{Parameters}&\multicolumn{1}{|c|}{\textbf{Values}} \\
			\cline{1-2} 
			\hline
			update{\textunderscore}frequency & 10.0\\
			static{\textunderscore}map & true\\
			\hline
		\end{tabular}
	\end{center}
\end{table}

\begin{table}[htbp]
	\caption{Local costmap parameters}
	\label{tab:table8}
	\begin{center}
		\begin{tabular}{|c|c|}
			\hline
			\textbf{Parameters}&\multicolumn{1}{|c|}{\textbf{Values}} \\
			\cline{1-2} 
			\hline
			update{\textunderscore}frequency & 10.0\\
			publish{\textunderscore}frequency & 10.0\\
			static{\textunderscore}map & false\\
			rolling{\textunderscore}window & true\\
			width & 6.0\\
			height & 6.0\\
			resolution & 0.05\\
			\hline
		\end{tabular}
	\end{center}
\end{table}

\section{Implementation of Photo taking behavior}
\label{Section:sec4}
The photo-taking behavior is implemented based on the Goal Controller node (\autoref{fig:fig14}). Instead of a user-defined goal position, the position of the detected face is used as a goal position for the robot to traverse to.

In this section, the implementation of two additional packages will be discussed: 1. Face Detector and 2. Robot Behavior.

\subsection{Face Detector package}
\label{Section:sec4a}
\begin{figure}[!htb]
	\centering
	\includegraphics[width=1.0\linewidth]{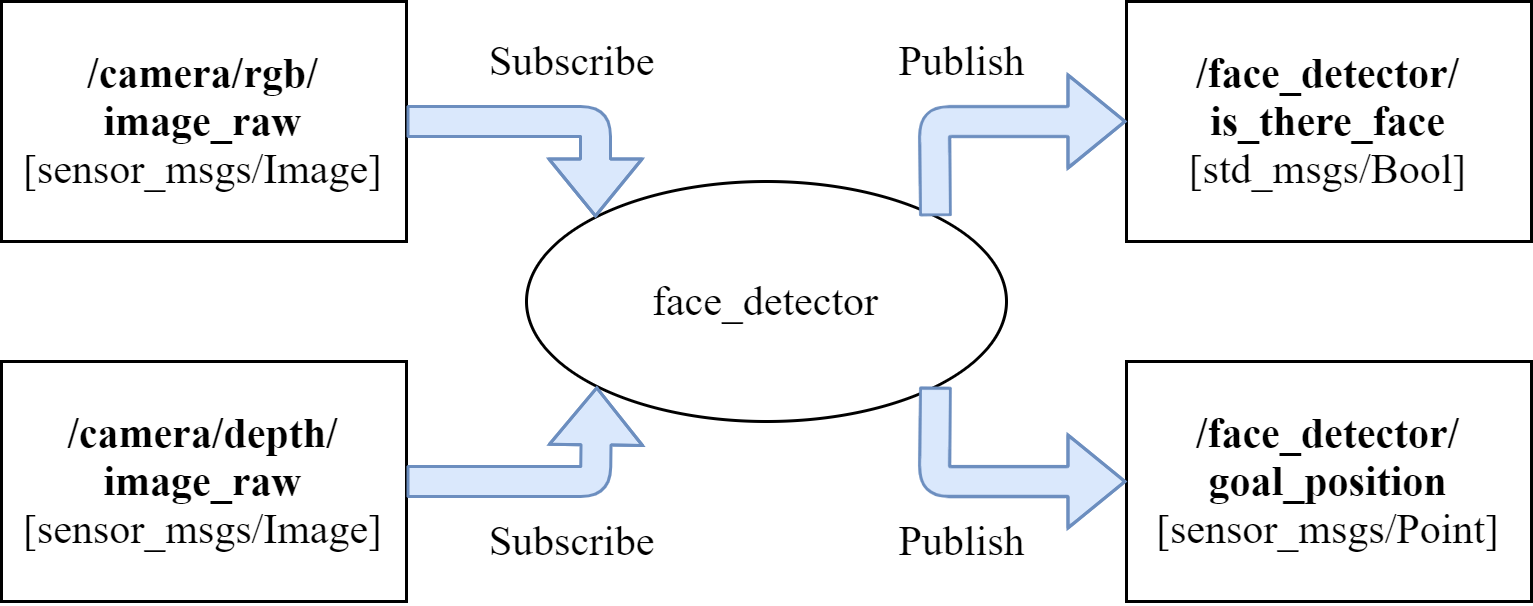}
	\caption{robot{\textunderscore}face{\textunderscore}detector package node}
	\label{fig:fig15}
\end{figure}

A package called \textbf{robot{\textunderscore}face{\textunderscore}detector} has been created which performs face detection on Kinect sensor RGB images and outputs the detected face's goal location based on the Depth image. Within the package, a node called \textit{face{\textunderscore}detector} which subscribes to the topics ``/camera/rgb/image{\textunderscore}raw" and ``/camera/depth/image{\textunderscore}raw" that contain messages published by \textbf{openni{\textunderscore}launch} node. As the names imply, these messages contains the RGB and Depth images information from the Kinect sensor. The \textit{face{\textunderscore}detector} node is shown in \autoref{fig:fig15}.

OpenCV Haar Cascade classifier \cite{opencv_library} has been used to detect human faces on the RGB image and the 2D location of the detected face is obtained from the face point cloud information derived from the depth image \cite{kamarudin2013method}. The OpenCV Bridge \cite{patrickjames} was used to convert the raw image message data to the cv2 image.

\subsection{Robot behavior configuration}
\begin{figure}[!htb]
	\centering
	\includegraphics[width=0.8\linewidth]{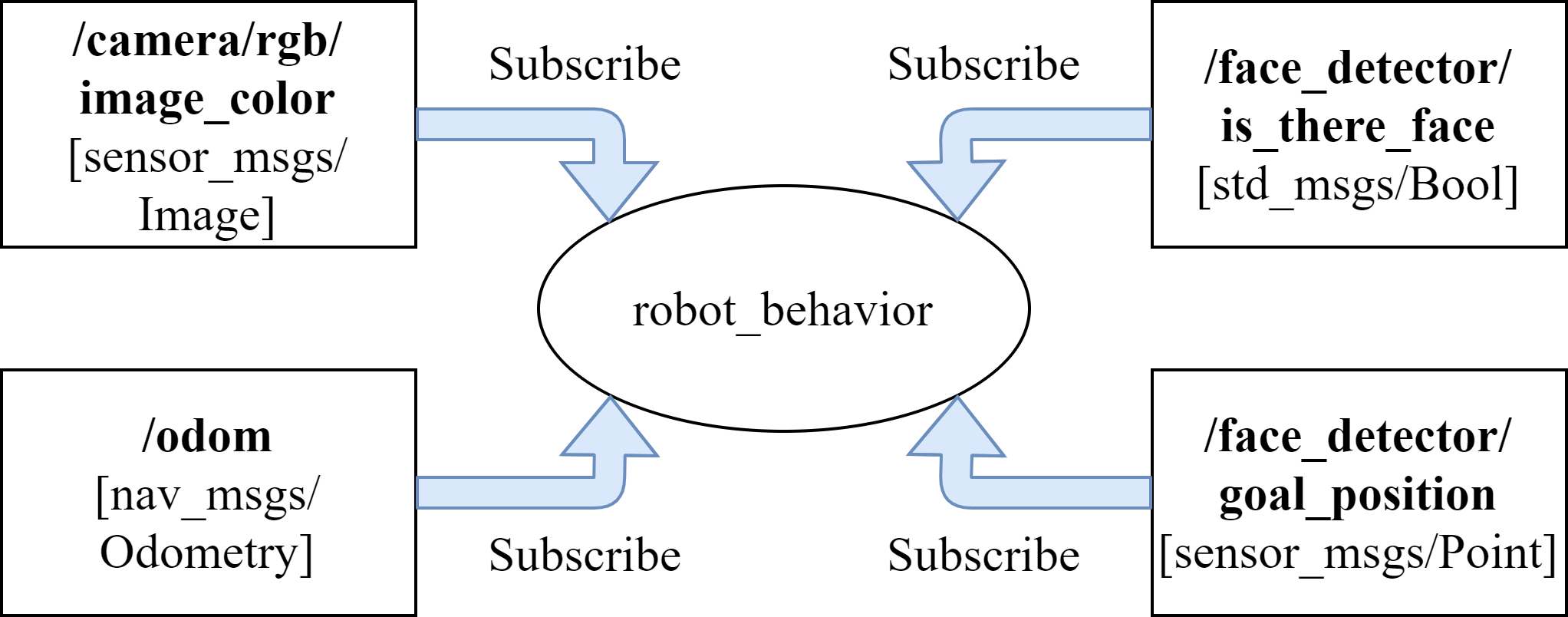}
	\caption{robot{\textunderscore}behavior package node}
	\label{fig:fig16}
\end{figure}

A package called \textbf{robot{\textunderscore}behavior} has been created which contains the behavior of the robot during exploration. Within the package, a node called \textit{robot{\textunderscore}behavior} shown in \autoref{fig:fig16} subscribes to the following four topics: 

\begin{itemize}
	\item ``/odom" which publishes messages of type ``nav{\textunderscore}msgs/Odometry"
	\item ``/camera/rgb/image{\textunderscore}color" which publishes messages of type ``sensor{\textunderscore}msgs/Image"
	\item ``/face{\textunderscore}detector/is{\textunderscore}there{\textunderscore}face" which publishes messages of type ``std{\textunderscore}msgs/Bool"
	\item ``/face{\textunderscore}detector/goal{\textunderscore}position" which publishes messages of type ``sensor{\textunderscore}msgs/Point"
\end{itemize}

The ``/odom" topic is to obtain the robot's current location so that when a face is detected, the position and orientation information that is needed to get the robot to the goal location (i.e face) can be computed relative to the current location of the robot. Then, the ``/face{\textunderscore}detector/is{\textunderscore}there{\textunderscore}face" topic and ``/face{\textunderscore}detector/goal{\textunderscore}position" topics are used in the behavior program for sending the location of the detected face to the Navigation stack using a ROS package called Actionlib \cite{actionlib}. Finally, the ``/camera/rgb/image{\textunderscore}color" topic is used to obtain the detected face images for saving.

By default, the behavior of the robot is to explore and roam its environment, and this functionality is provided by the Explorer package \cite{jim} mentioned earlier in \autoref{Section:sec3d}. When the face{\textunderscore}detector (\autoref{Section:sec4a}) package has detected a face, the robot will start navigating toward the face location to initiate taking a picture by using OpenCV's save() and show() function \cite{opencv_library}.

\section{Experiment setup}
\begin{figure}[!htb]
	\centering
	\includegraphics[width=0.8\linewidth]{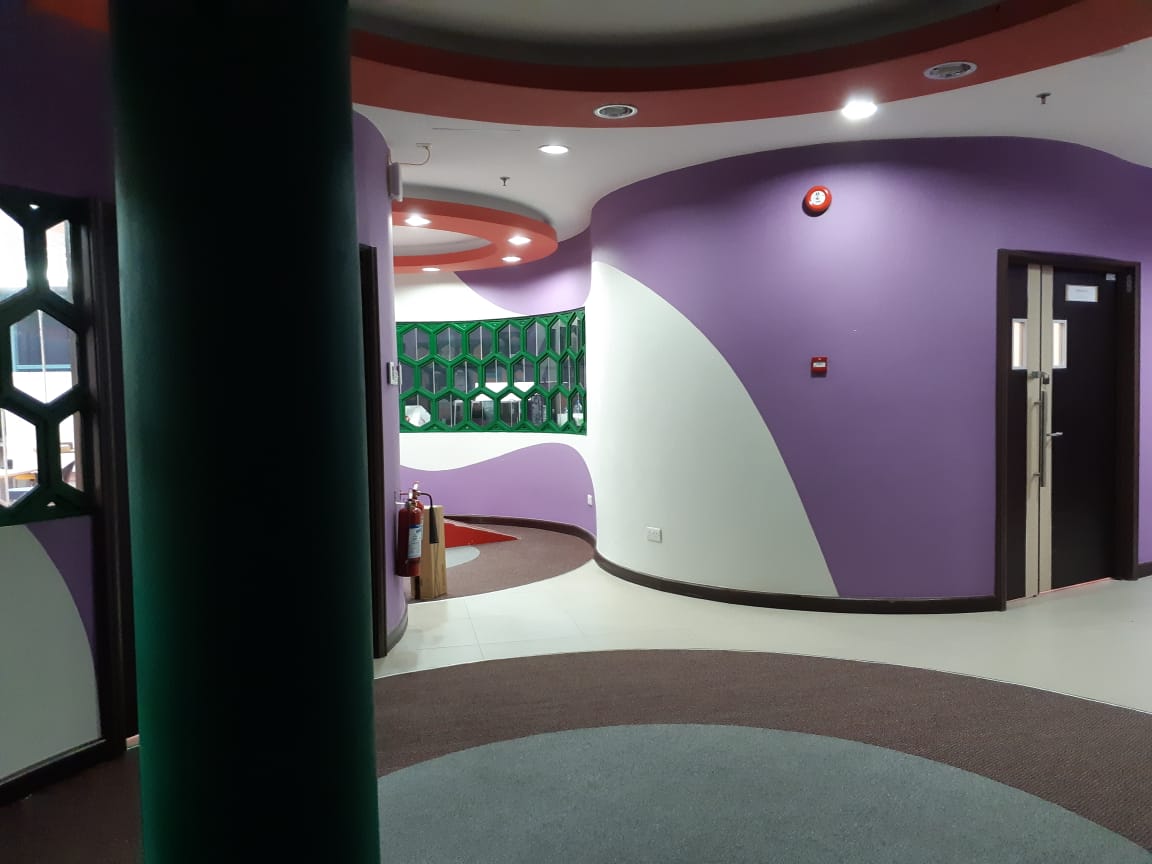}
	\caption{Outside environment of working lab (angle A)}
	\label{fig:fig17}
\end{figure}

\begin{figure}[!htb]
	\centering
	\includegraphics[width=0.8\linewidth]{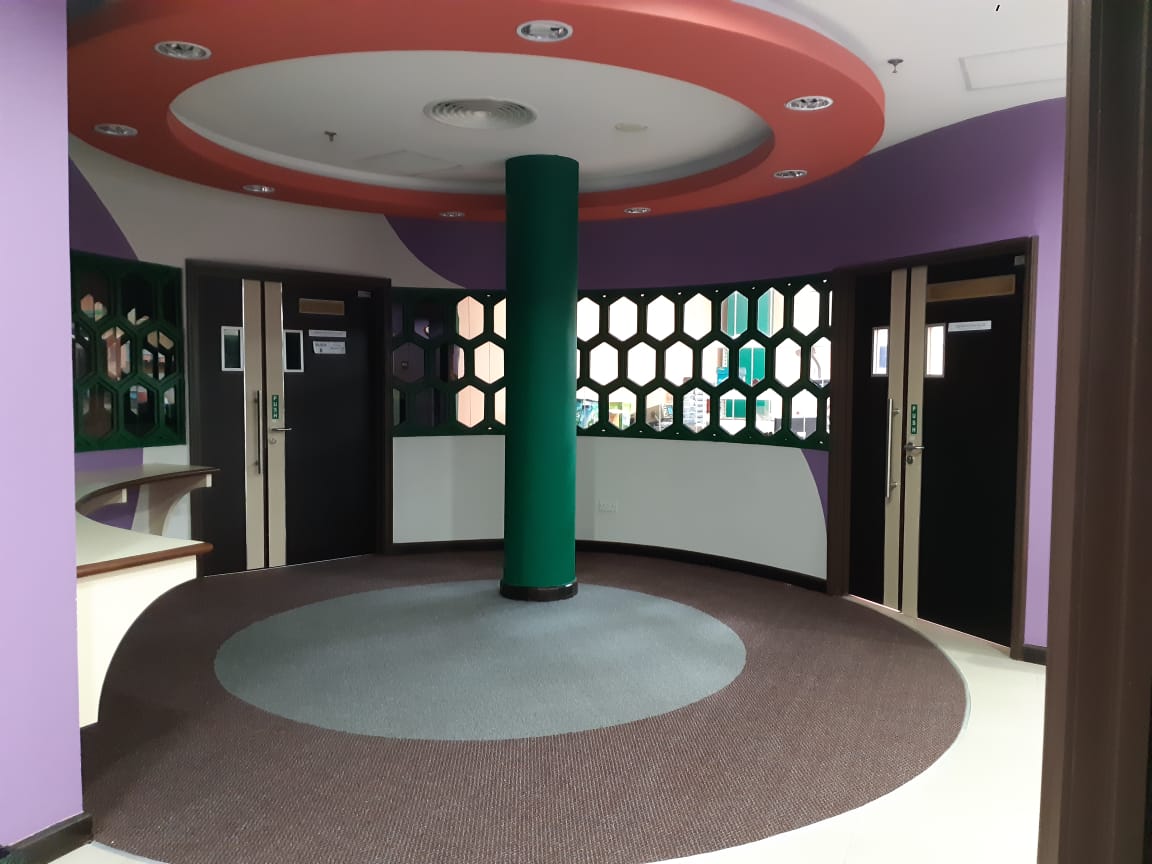}
	\caption{Outside environment of working lab (angle B)}
	\label{fig:fig18}
\end{figure}

The evaluation of the implemented functionalities (i.e navigation system and photo-taking behavior) was carried out in an indoor environment shown in \autoref{fig:fig17} and \autoref{fig:fig18}. The confirmation of results was through visual inspections due to the lack of a mocap system.

The evaluation was performed on the following functions:

\begin{enumerate}
	\item \textbf{Wheel encoder odometry}: a test to determine correct implementation of encoder-based odometry.
	\item \textbf{Mapping and localization}: a test to determine correct implementation of encoder-based odometry with SLAM.
	\item \textbf{Autonomous navigation}: capability to reach a goal location while avoiding obstacle.
	\item \textbf{Face detection}: capability to detect face and compute its location in 2D map.
	\item \textbf{Photo taking behavior}: capability to use the implemented navigation system to traverse to a human reliably and take photo.
\end{enumerate}

\autoref{tab:table9} shows the set travel speed and acceleration of the robot during the experiment. These are also parameters of the Navigation stack similar to \autoref{tab:table5}, \autoref{tab:table6}, \autoref{tab:table7} and \autoref{tab:table8}. During the experiment, visual inspection in addition to RVIZ were used to monitor the robot. At the beginning of each experiment, the robot was moved to its starting position, and the necessary commands to run the functions for evaluation listed above were sent through a remote computer.
\begin{table}[htbp]
	\caption{Robot travel speed and acceleration setup}
	\label{tab:table9}
	\begin{center}
		\begin{tabular}{|c|c|}
			\hline
			\textbf{Parameters}&\multicolumn{1}{|c|}{\textbf{Values}} \\
			\cline{1-2} 
			\hline
			max{\textunderscore}vel{\textunderscore}x & 0.175 m/s\\
			min{\textunderscore}vel{\textunderscore}x & 0.175 m/s\\
			max{\textunderscore}vel{\textunderscore}theta & 1.0 m/s\\
			min{\textunderscore}vel{\textunderscore}theta & -1.0 m/s\\
			min{\textunderscore}in{\textunderscore}place{\textunderscore}velocity{\textunderscore}theta & 1.0 m/s\\
			escape{\textunderscore}vel & -0.175 m/s\\
			acc{\textunderscore}lim{\textunderscore}theta & 0.5 m/s\\
			acc{\textunderscore}lim{\textunderscore}x & 0.05 m/s\\
			acc{\textunderscore}lim{\textunderscore}y & 0.05 m/s\\
			\hline
		\end{tabular}
	\end{center}
\end{table}

The maximum and minimum velocity in x direction (max{\textunderscore}vel{\textunderscore}x, min{\textunderscore}vel{\textunderscore}x) were set at 0.175 m/s for fixed speed navigation. Similarly, the maximum and minimum rotational velocity (max{\textunderscore}vel{\textunderscore}theta, min{\textunderscore}vel{\textunderscore}theta) were set at 1.0 m/s and -1.0 m/s respectively for fixed speed rotation during navigation. The minimum in-place rotational velocity (min{\textunderscore}in{\textunderscore}place{\textunderscore}vel{\textunderscore}theta) was set at 1.0 m/s. The escape velocity (escape{\textunderscore}vel) which is the reverse speed was set at -0.175 m/s. The acceleration of velocity in x and y direction (acc{\textunderscore}lim{\textunderscore}x, acc{\textunderscore}lim{\textunderscore}y) was set at 0.05 m/s\textsuperscript{2} to ensure smooth navigation and no jerkiness. Similarly, the maximum acceleration of rotational velocity (acc{\textunderscore}lim{\textunderscore}theta) was set at 0.5 m/s\textsuperscript{2} for smooth rotation during navigation.

\section{Results}	

\subsection{Wheel encoder odometry}
\label{Section:sec6a}
In this experiment, the Base Controller and Teleoperation package (\autoref{Section:sec3d}) were used to manually control the robot from a remote computer. In addition, Odometry source (\autoref{Section:sec3d}) was used to compute the robot's encoder Odometry. We remotely controlled the robot and through visual inspection, the robot's actual location correlated with the estimated Odometry readings.

\subsection{Mapping and localization}
SLAM package (\autoref{Section:sec3d}) was used to build a map of the perceived environment. We remotely controlled the robot and through the RVIZ interface, the generated map correlated with the actual test environment shown in \autoref{fig:fig16} and \autoref{fig:fig17}. Additionally, the robot's position throughout the test correlated with the estimated Odometry readings.

\begin{figure}[!htb]
	\centering
	\includegraphics[width=0.8\linewidth]{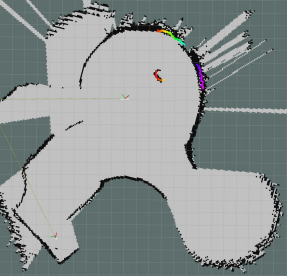}
	\caption{Generated map from slam{\textunderscore}mapping node}
	\label{fig:fig19}
\end{figure}

\subsection{Autonomous navigation}
\label{Section:sec6c}
In this experiment, all packages that were shown in \autoref{tab:table4} were run and an obstacle in the form of cardboard boxes was laid in the middle of the test environment. Through the RVIZ interface, a goal location was sent and the robot started its navigation and stopped at the set goal location without collisions. Moreover, the robot traversed according to the trajectory generated by the Navigation stack and moved at a smooth and steady pace. A drawback with using the Navigation stack is that it requires fine-tuning in the settings in order for it to work properly with different robots and this can be a time-consuming trial and error process. Additionally, the Navigation stack will not run smoothly on an embedded system like the Raspberry Pi and would require setting up an additional computer with more computational power to run the autonomous navigation.

\begin{figure}[!htb]
	\centering
	\includegraphics[width=0.8\linewidth]{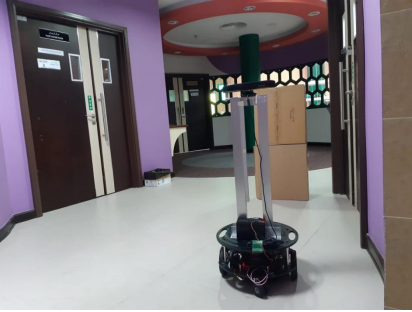}
	\caption{Environment with added obstacle}
	\label{fig:fig20}
\end{figure}

\begin{figure}[!htb]
	\centering
	\includegraphics[width=0.8\linewidth]{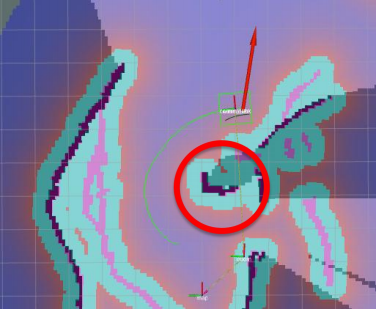}
	\caption{Global path trajectory of robot goal position}
	\label{fig:fig21}
\end{figure}

\subsection{Face detection}
\label{Section:sec6d}
In this experiment, a human subject sat in front of the robot at varying distances and the robot can reliably detect human faces at a distance of up to 3.0 meters. The face point cloud that was computed correlated to the actual position of the detected face in the test environment. However, the OpenCV Haar cascade \cite{opencv_library} face detection implementation used in this paper was prone to false positives and sometimes detect faces in a location where there are were no faces. The face detector is capable of detecting frontal faces. However, humans may not always be in a frontal position or within a short distance of 3.0 meters (\autoref{Section:sec6d}). So, human detection can be implemented as a prior before initiating face detection for the robot's photo-taking behavior.

\begin{figure}[!htb]
	\centering
	\includegraphics[width=0.8\linewidth]{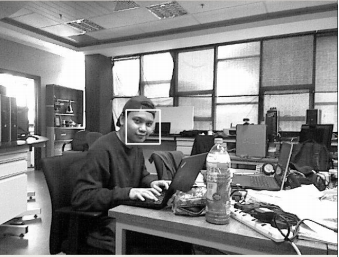}
	\caption{Detected human face at 1.0 meter away}
	\label{fig:fig22}
\end{figure}

\begin{figure}[!htb]
	\centering
	\includegraphics[width=0.8\linewidth]{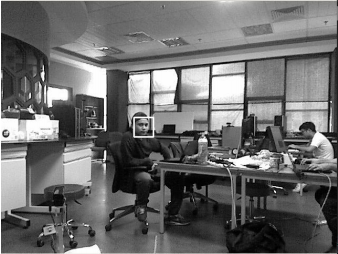}
	\caption{Detected human face at 3.0 meter away}
	\label{fig:fig23}
\end{figure}

\subsection{Photo taking behavior}
During the experiment, the robot managed to navigate to the human subject using the implemented navigation system for photo taking, subsequently rotating and continue exploration until another human face is detected. There were instances when the face detector failed to correctly detect faces and this resulted in the robot navigating to a location and a taking a picture with no humans.

\section{Conclusion}
In this paper, we have described the implementation of ROS Navigation stack on a Parallax Eddie mobile robot platform which was customized and fitted with different hardware components. We have given detailed information of the robot hardware specifications which included the wiring diagram and a list of hardware used for the robot. In terms of software contributions, we have created several custom ROS packages for the Navigation stack such as robot odometry, goal controller, base controller, teleoperation and sensor transforms.  Additionally, we have shown a detailed implementation of the robot's motor control with Pulse Width Modulation (PWM) and wheel encoder-based odometry localization. Furthermore, the parameters of the Navigation stack have been tuned according to the robot's hardware specifications. Finally, two more ROS packages were implemented which are face detector and photo-taking behavior to demonstrate a use case for autonomous navigation.

An important component that needs to be improved is that the face detector package used in the experiments cannot reliably detect human faces in non-frontal positions and from a distance of more than 3.0 meters. So, perhaps a more robust face detection feature that incorporates human detection as a prior before face detection can be implemented using Kinect sensor’s capability to detect skeletal joints of human bodies. This will allow the robot to know the orientation of the human and allow it to traverse to the front side of the human and subsequently allow the face detector to perform optimally since the robot will be facing the front side of a human and be within the range of 3.0 meters. 

A GitLab repository containing details of the experimental setup as well as source codes used for the robot system has been made public in \cite{ailab}.

\printbibliography

\end{document}